\documentclass[runningheads]{llncs}
\usepackage[T1]{fontenc}
\usepackage{graphicx}
\usepackage{graphicx}
\usepackage{booktabs}
\usepackage{tabularx}
\usepackage{multirow}
\usepackage{amsmath,amssymb}
\graphicspath{{fig/}}

\newcommand{\cofirst}{\textsuperscript{\textdagger}}
\newcommand{\corrauth}{\textsuperscript{*}}
\usepackage{xcolor}
\usepackage[colorlinks=true, linkcolor=black, citecolor=black, urlcolor=blue]{hyperref}

\begin{document}
\title{Transition-Based Digital Twin Modelling for Alzheimer's Disease under Sparse Longitudinal Data}
%
%
\titlerunning{A Transition-Based Digital Twin Framework for AD Prediction}

\author{
Yinyu Huang\cofirst\inst{1}
\and
Yilin Zhang\cofirst\inst{1}
\and
Sofia Michopoulou\inst{2,3}
\and
Christopher Kipps\inst{2,3}
\and
Rahman Attar\corrauth\inst{1}
}

\authorrunning{Y. Huang and Y. Zhang et al.}

\institute{
School of Electronics and Computer Science, University of Southampton, UK
\and
University Hospital Southampton NHS Foundation Trust, Southampton, UK
\and
Faculty of Medicine, University of Southampton, UK\\[1mm]
}
\maketitle              
\begingroup
\renewcommand{\thefootnote}{\textdagger}
\footnotetext{These authors contributed equally to this work.}
\renewcommand{\thefootnote}{*}
\footnotetext{Corresponding author: r.attar@southampton.ac.uk}
\endgroup
\begin{abstract}
Alzheimer’s disease (AD) progression is highly heterogeneous and is typically observed through sparse and irregularly sampled longitudinal data, posing challenges for reliable prediction and individualised monitoring. Existing machine learning approaches have improved AD prediction using multimodal data, yet they often focus on static classification or cohort-level risk estimation, providing limited support for subject-specific trajectory modelling and uncertainty-aware reasoning.
To address these limitations, we present a personalised digital twin framework for Alzheimer’s disease prediction and scenario-based analysis using multimodal longitudinal data. The proposed approach integrates complementary modelling strategies to capture both short-term clinical transitions and longer-range temporal dependencies across follow-up visits. Using data from the Alzheimer’s Disease Neuroimaging Initiative (ADNI), including cognitive assessments, clinical variables, and MRI-derived structural phenotypes, the framework predicts future cognitive status and diagnostic categories while quantifying predictive uncertainty and enabling patient-specific what-if trajectory analysis under controlled perturbations of clinically meaningful variables.
Experimental evaluation on leak-free subject-level splits demonstrates strong performance in both cognitive score forecasting and diagnosis classification. Importantly, in this sparse and irregular ADNI setting, transition-based modelling of adjacent visits achieved higher predictive accuracy than the sequence-based branch, suggesting that local transition modelling may be more data-efficient under the present data constraints. While sequence models remain valuable for uncertainty-aware trajectory forecasting, local transition modelling offers a more data-efficient and robust predictive strategy in this setting.
These findings highlight the importance of aligning temporal modelling strategies with the structure of real-world clinical data and suggest that transition-based digital twin formulations may provide a practical and interpretable approach for personalised disease forecasting and monitoring in neurodegenerative disorders.

\end{abstract}
\section{Introduction}
Alzheimer’s disease (AD) is a major public health challenge and a leading cause of cognitive decline in older adults \cite{norton2014potential}. Clinical progression is heterogeneous across individuals and is often observed through sparse, irregularly sampled follow-up. Although multimodal machine learning has improved AD prediction by integrating structural MRI, cognitive assessments, and clinical variables, many existing studies remain focused on static classification or cohort-level risk estimation rather than subject-specific longitudinal trajectories \cite{odusami2024machine,qiang2023diagnosis,hasan2024explainable,kumar2024himal,karaman2024assessing}. As a result, they provide limited support for individualised monitoring and prospective reasoning.

Several gaps remain in longitudinal AD modelling. First, many methods emphasise point prediction without uncertainty quantification, which limits their ability to extrapolate from incomplete clinical histories \cite{marinescu2020alzheimer,fisher2019machine}. Second, even when time-series models are used, the outputs usually remain at a benchmark level rather than providing subject-level predictive representations that can be updated with available observations and queried for scenario-based analysis. Third, what-if analyses are rarely integrated into the same framework, despite their potential value for exploring how projected disease trajectories may change under hypothetical variations in clinically meaningful covariates.

Digital twin concepts offer a natural way to address these limitations. In healthcare, a digital twin can be viewed as a virtual representation of an individual that is updated with observed data and used for personalised prediction, monitoring, and scenario analysis \cite{balasubramaniam2024machine,ren2025utilization,elgammal2025digital}. However, most medical digital twin work remains conceptual, single-modality, or targeted at domains other than multimodal longitudinal neurodegeneration \cite{amato2025digital,fekonja2024digital}. This motivates a more practical AD-oriented formulation that supports both subject-level forecasting and uncertainty-aware what-if trajectory analysis. A further challenge is that sparse longitudinal cohorts often favour different modelling strategies depending on prediction horizon: transition-based models can exploit many local training pairs, whereas sequence models may better reflect temporal continuity but require stronger assumptions about missingness and alignment. In this context, we define a digital twin as a subject-specific predictive representation that can be updated with available longitudinal observations and queried for forecasting and scenario-based what-if analysis.

In this work, we propose a personalised digital twin framework for Alzheimer’s disease using multimodal longitudinal data. The framework combines a transition-based branch for short-horizon prediction with a sequence-based branch for longer-range forecasting and uncertainty-aware scenario analysis. Beyond the framework itself, the main insight of this study is that, in sparse and irregular clinical follow-up settings, modelling local transitions between adjacent visits can be more robust and practically useful than relying solely on sequence-based temporal models. The main contributions of this work are threefold. First, we present a hybrid digital twin framework for individualised prediction of cognitive decline and diagnosis in Alzheimer’s disease. Second, we show empirically that a transition-based modelling strategy can outperform a sequence-based strategy under sparse longitudinal clinical data conditions. Third, we demonstrate how the framework supports uncertainty-aware forecasting and patient-specific what-if simulation.

The code developed for this study is publicly available at \url{https://github.com/YilinZhang00/Digital-Twin-for-AD.git}.

\section{Methods}
\subsection{Dataset and preprocessing}
We used data from the Alzheimer’s Disease Neuroimaging Initiative (ADNI), integrating clinical assessments, cognitive scores, and FreeSurfer-derived MRI structural phenotypes by subject identifier (RID). After harmonisation, the working dataset contained 2801 records across 760 unique subjects with six-month intervals from baseline to 60 months. MMSE was treated as a continuous target and diagnosis (DX) as a three-class target: cognitively normal (CN), mild cognitive impairment (MCI), and Alzheimer’s disease (AD). After removing identifiers and targets, 385 predictors remained.

The feature set included both static and dynamic variables. Static variables were features that remained constant across follow-up, such as sex, education, APOE genotype, intracranial volume, and baseline cognitive descriptors. Dynamic variables included longitudinal cognitive measures and MRI-derived regional phenotypes. Categorical variables were numerically encoded and numerical features were z-score normalised. Follow-up was aligned to a six-month grid (baseline, m06, m12, \ldots, m60). For irregularly sampled individuals, missing values in the temporal branch were linearly interpolated after train-only preprocessing. Linear interpolation was used as a simple and transparent strategy to align irregular follow-up for the sequence-based temporal branch. However, this preprocessing step introduces artificial regularity into the longitudinal trajectories, because it creates equally spaced observations that may not fully reflect true clinical follow-up patterns. Consequently, the subsequent comparison between the two branches should be interpreted as providing an upper-bound estimate of sequence model performance under idealised regularity, rather than a definitive statement about their relative merits in fully irregular settings. Subjects with substantial missingness at critical time points were excluded to ensure minimum longitudinal consistency for modelling, resulting in 760 participants. This reflects a common trade-off in longitudinal clinical studies between data completeness and cohort size.

\subsection{Hybrid digital twin design}
\begin{figure}[t]
\centering
\includegraphics[width=0.98\textwidth]{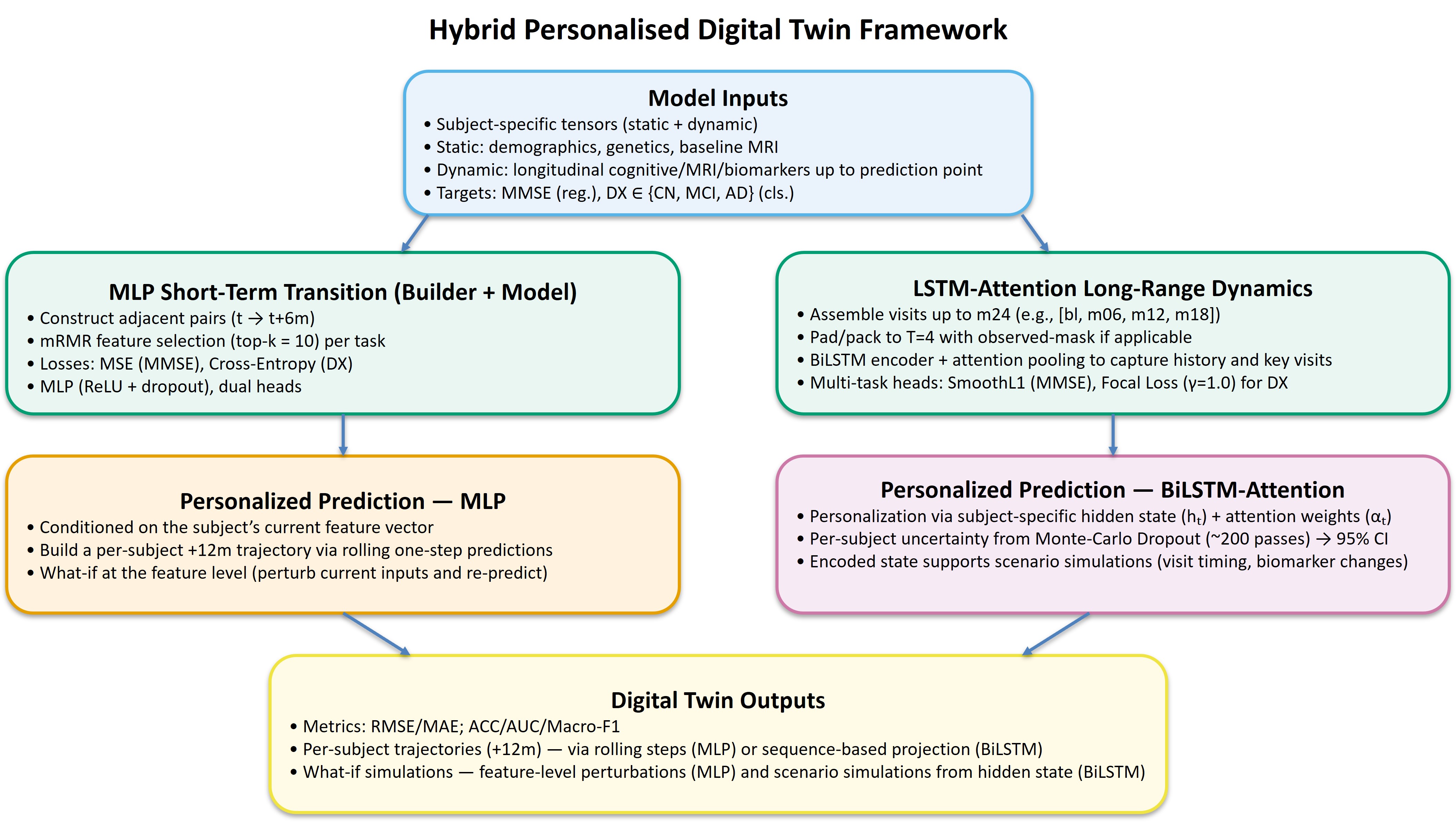}
\caption{Architecture of the hybrid personalised digital twin framework. The MLP branch models adjacent short-term transitions, while the BiLSTM-Attention branch captures longer-range longitudinal dependencies and supports uncertainty-aware forecasting and what-if trajectory analysis.}
\label{fig:framework}
\end{figure}

The framework is shown in Fig.~\ref{fig:framework}. The short-horizon branch of the digital twin framework is an MLP-based transition model that operates on adjacent visit pairs $(t \rightarrow t{+}6\text{ months})$. For each pair, the input contains the current visit’s clinical, cognitive, and imaging features, and the targets are the next visit’s MMSE and DX. This design increases the effective sample size and aligns the task with local state transitions. The transition-based formulation is motivated by the observation that sparse longitudinal clinical data often contain more reliable local transition signals than fully aligned long-range temporal structure. The MLP models for MMSE regression and DX classification are trained separately using mRMR-selected features, Adam optimisation, and early stopping.

The longer-range branch of the digital twin framework is a BiLSTM-Attention sequence model designed to capture temporal dependencies across aligned follow-up visits. For this branch, each subject’s follow-up was aligned to baseline, m06, m12, m18, and m24. The model uses observations up to m18 to predict the 24-month endpoint, with attention pooling used to summarise the sequence representation. The regression head predicts MMSE with SmoothL1 loss, while the classification head predicts DX with focal loss to address class imbalance. For uncertainty-aware forecasting, dropout remains active at test time and repeated stochastic passes are used to form predictive intervals. For what-if analyses, clinically meaningful covariates are perturbed at the forecast origin while all remaining inputs are fixed. The two branches are designed to serve complementary purposes: the transition-based MLP focuses on robust short-horizon predictive performance, whereas the BiLSTM-Attention branch emphasises longer-range temporal continuity, predictive uncertainty, and subject-specific what-if trajectory analysis.

\subsection{Experimental setup}
We used subject-level leak-free splits of 70\%/20\%/10\% for train/validation/test, corresponding to 532, 152, and 76 subjects. All preprocessing transformations were fitted on the training set only and then applied to validation and test sets. For the MLP, adjacent visit pairing yielded approximately 2005 samples. For the BiLSTM-Attention model, we retained only subjects with aligned baseline-to-m24 trajectories, at least two actual dynamic observations among m06/m12/m18, and available labels at m24, resulting in 696 usable sequence samples. The two branches are compared as alternative digital twin formulations under realistic data usage conditions rather than as perfectly horizon-matched architectures; accordingly, the comparison is intended to highlight practical trade-offs between local transition modelling and longer-horizon temporal forecasting.

For MLP feature compression, we used minimum Redundancy Maximum Relevance (mRMR) on the training split, with the number of selected features $K$ chosen on validation data and then fixed for final evaluation. We report RMSE and MAE for MMSE regression, and ACC, macro-averaged AUC, and macro-F1 for diagnosis classification. Unless otherwise stated, confidence intervals are obtained by subject-level bootstrapping on the held-out test set.

\section{Results}
\subsection{Main model comparison}
Table~\ref{tab:main} summarises the held-out test performance of the two digital twin branches. The MLP branch consistently outperformed the BiLSTM-Attention branch across both regression and classification metrics. In the current data setting, the MLP achieves RMSE 2.149 and MAE 1.529 for MMSE prediction, while also reaching ACC 0.906, AUC 0.976, and macro-F1 0.908 for diagnosis classification. The BiLSTM-Attention branch remains competitive but weaker, with RMSE 2.687, MAE 1.926, ACC 0.806, AUC 0.928, and macro-F1 0.798. This gap is consistent with the different sample construction strategies: the MLP benefits from many local time pairs, whereas the sequence model learns from fewer aligned trajectories and performs a more challenging longer-horizon forecast. These results suggest that, under sparse and partially irregular longitudinal follow-up, modelling short-term local transitions may be a more data-efficient strategy than learning longer-range sequence representations. Although the temporal branch underperformed on point prediction metrics, it remains valuable for individualised uncertainty-aware forecasting and scenario-based trajectory exploration, which are central properties of a digital twin framework.

\begin{table}[t]
\caption{Held-out test performance comparison between the two digital twin branches.}
\label{tab:main}
\centering
\begin{tabular}{lccccc}
\toprule
Model & RMSE & MAE & ACC & AUC & Macro-F1\\
\midrule
MLP & 2.149 & 1.529 & 0.906 & 0.976 & 0.908\\
BiLSTM-Attention & 2.687 & 1.926 & 0.806 & 0.928 & 0.798\\
\bottomrule
\end{tabular}
\end{table}

\subsection{Validation analyses for the MLP branch}
The MLP branch was further examined on the validation split to understand the role of feature selection, feature-source composition, and conventional classifier baselines. We first scanned the number of retained mRMR features and then analysed the effect of removing feature selection or restricting the input to static-only or dynamic-only subsets. Finally, the selected MLP configuration was compared with multinomial logistic regression, random forest, and decision tree classifiers under the same reduced feature space.

\begin{table}[t]
\caption{Validation analyses under the shared preprocessing pipeline.}
\label{tab:valscan}
\centering
\small
\begin{tabular}{lccccc}
\toprule
\multicolumn{6}{l}{\textbf{(a) mRMR feature-count scan (validation)}}\\
\midrule
$K$ & ACC & Macro-F1 & AUC & RMSE & MAE\\
5  & 0.929 & 0.929 & 0.979 & 2.204 & 1.651\\
10 & 0.936 & 0.936 & 0.984 & 2.157 & 1.619\\
15 & 0.943 & 0.943 & 0.984 & 2.153 & 1.606\\
20 & 0.943 & 0.943 & 0.985 & 2.263 & 1.648\\
30 & 0.926 & 0.926 & 0.983 & 2.550 & 1.874\\
\midrule
\multicolumn{6}{l}{\textbf{(b) Ablations (validation)}}\\
\midrule
Setting & RMSE & MAE & ACC & AUC & Macro-F1\\
MLP + mRMR         & 2.154 & 1.612 & 0.943 & 0.984 & 0.942\\
MLP w/o mRMR       & 4.261 & 3.350 & 0.762 & 0.923 & 0.758\\
MLP (static-only)  & 3.021 & 2.222 & 0.782 & 0.920 & 0.781\\
MLP (dynamic-only) & 4.584 & 3.597 & 0.712 & 0.890 & 0.708\\
\bottomrule
\end{tabular}
\end{table}

\begin{table}[t]
\caption{DX baselines on the validation split using the shared mRMR feature space ($K{=}15$).}
\label{tab:dxbase}
\centering
\begin{tabular}{lccc}
\toprule
Model & ACC & AUC & Macro-F1\\
\midrule
MLP (ours)          & 0.943 & 0.984 & 0.943\\
Logistic Regression & 0.936 & 0.983 & 0.936\\
Random Forest       & 0.922 & 0.980 & 0.922\\
Decision Tree       & 0.880 & 0.955 & 0.881\\
\bottomrule
\end{tabular}
\end{table}

Tables~\ref{tab:valscan} and \ref{tab:dxbase} show that mRMR markedly improves both regression and classification relative to using all input features without selection, and that the static-dynamic combined setting is clearly stronger than using dynamic features alone. Logistic regression remains highly competitive in the reduced feature space, indicating that much of the classification gain comes from informative and non-redundant feature selection rather than purely architectural complexity. The competitiveness of logistic regression in the selected feature space further supports the view that the main benefit arises from informative and non-redundant longitudinal representation, rather than from nonlinear classifier complexity per se. Across the feature-count scan, $K{=}15$ provided the best balance between predictive performance and parsimony. These findings indicate that predictive performance is driven largely by the quality and relevance of the reduced feature representation, rather than by model complexity alone. In this setting, careful feature selection and local transition modelling appear more important than adopting deeper temporal architectures.

\subsection{MLP branch test visualisation and calibration}
\begin{figure}[t]
\centering
\begin{minipage}[t]{0.48\textwidth}
\centering
\includegraphics[width=\linewidth]{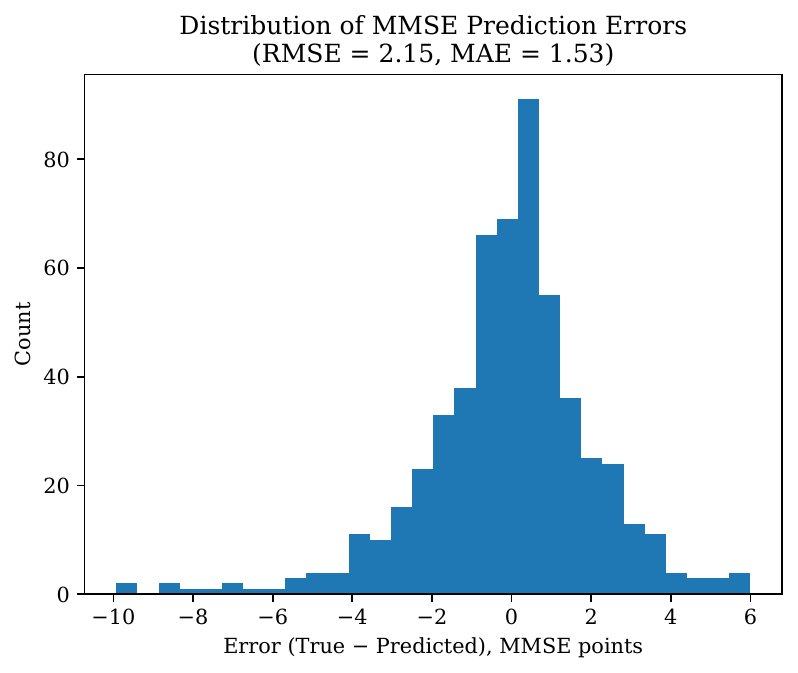}\\[-0.2em]
{\small (a) MMSE error distribution.}
\end{minipage}\hfill
\begin{minipage}[t]{0.48\textwidth}
\centering
\includegraphics[width=\linewidth]{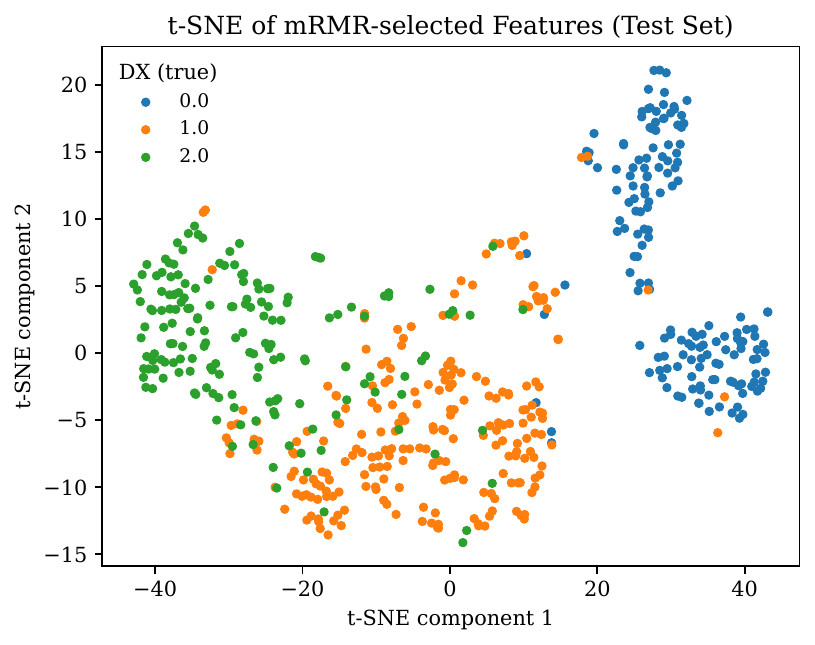}\\[-0.2em]
{\small (b) t-SNE of selected features for DX.}
\end{minipage}
\caption{MLP branch visualisation on the held-out test set.}
\label{fig:mlp}
\end{figure}

Fig.~\ref{fig:mlp} shows the MLP branch on the held-out test set. The MMSE error distribution is centred close to zero, with most errors falling within roughly $\pm 2$ points. The t-SNE visualisation of the mRMR-selected features shows clear separation among CN, MCI, and AD samples, consistent with the high AUC and macro-F1 reported in Table~\ref{tab:main}. These plots support the interpretation that the feature-reduced MLP branch captures a stable and discriminative representation for both tasks.

\begin{figure}[t]
\centering
\includegraphics[width=0.68\textwidth]{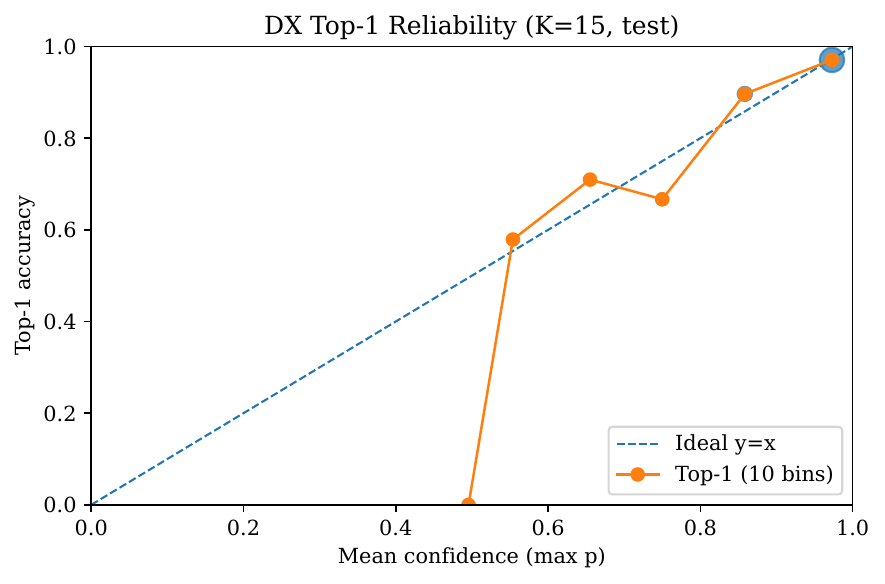}
\caption{Top-1 reliability plot for the MLP+mRMR diagnosis classifier on the held-out test set.}
\label{fig:calib}
\end{figure}

The chosen MLP with mRMR configuration also exhibited reasonable probability calibration at the Top-1 operating point. The multiclass Brier score was 0.1423, the overall $\mathrm{ECE}_{\mathrm{top1}}$ was 0.0185, and the macro class-wise ECE was 0.0177, indicating that predicted probabilities were broadly aligned with empirical accuracy. This is particularly important for future clinical decision support settings, where overconfident incorrect predictions may be more problematic than moderate reductions in raw accuracy. Fig.~\ref{fig:calib} shows the Top-1 reliability curve used for this analysis. The curve remains close to the identity line across most confidence bins, with only a modest deviation at the highest confidence range.

\subsection{BiLSTM-Attention branch}
\begin{figure}[t]
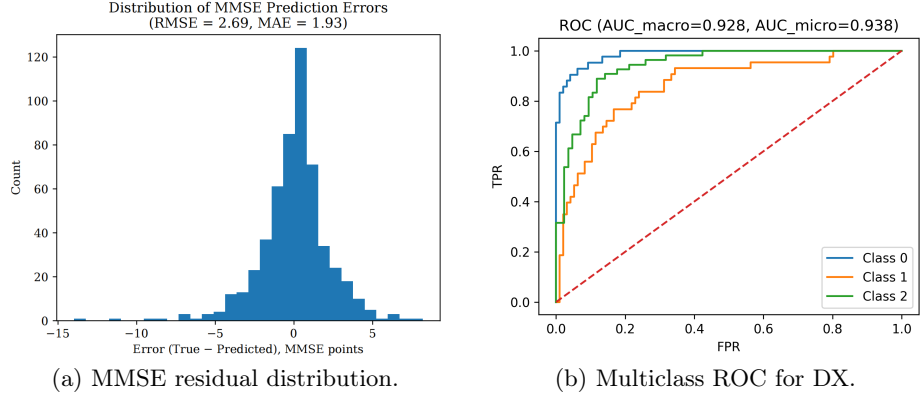

\centering
\begin{minipage}[t]{0.48\textwidth}
\centering
\includegraphics[width=\linewidth]{fig/fig04_mmse-residuals.pdf}\\[-0.2em]
{\small (a) MMSE residual distribution.}
\end{minipage}\hfill
\begin{minipage}[t]{0.48\textwidth}
\centering
\includegraphics[width=\linewidth]{fig/fig05_dx-roc.pdf}\\[-0.2em]
{\small (b) Multiclass ROC for DX.}
\end{minipage}
\caption{BiLSTM-Attention branch visualisation on the held-out test set.}
\label{fig:lstm}
\end{figure}

For the BiLSTM-Attention digital twin, residuals remained centred around zero, with most MMSE errors falling within roughly $\pm 3$ points. The diagnosis branch achieved macro-AUC 0.928 and macro-F1 0.798, demonstrating meaningful class discrimination even though it remained below the MLP. Its role in the framework is therefore not to serve as the strongest predictive baseline, but to provide a temporally coherent subject-level forecast with predictive intervals and what-if trajectory analysis capability. Fig.~\ref{fig:lstm} shows the residual distribution and the multiclass ROC curves for the temporal model. Although weaker quantitatively, this branch plays a distinct role in the overall framework because it supports uncertainty-aware multi-step forecasting and scenario-based trajectory analysis.

\subsection{Personalised Digital Twin Prediction and what-if trajectory analysis}

To illustrate the flexibility and individual-level interpretability of the proposed digital twin framework, three representative patient-level personas are examined, each reflecting a distinct longitudinal monitoring scenario in Alzheimer’s disease. RID=22 represents an idealised stable case with regular follow-up and coherent trajectories across modelling components. RID=223 represents a real-world progressive case with irregular follow-up and diagnostic drift from cognitively normal (CN) toward mild cognitive impairment (MCI) and Alzheimer’s disease (AD), despite comparatively stable MMSE. RID=898 represents a stable-appearing case, in which MMSE remains relatively high while model-estimated diagnostic risk shifts toward AD. Together, these personas illustrate complementary behaviours of the framework under different longitudinal conditions.

\begin{figure}[t]
\centering
\setlength{\tabcolsep}{4pt}
\renewcommand{\arraystretch}{1.0}
\begin{tabular}{@{}c c c@{}}

\begin{minipage}[c]{0.31\textwidth}
\centering
\includegraphics[width=\linewidth]{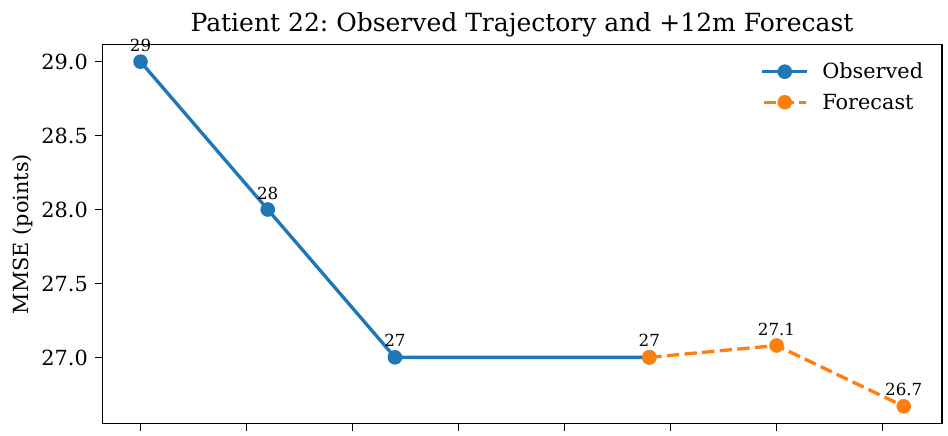}\\[-0.2em]
{\small (a) RID=22: MMSE}
\end{minipage} &
\begin{minipage}[c]{0.31\textwidth}
\centering
\includegraphics[width=\linewidth]{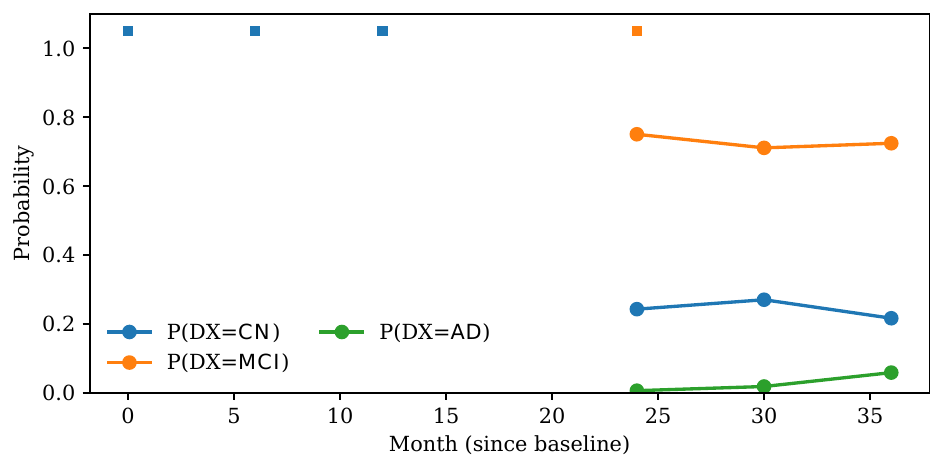}\\[-0.2em]
{\small (b) RID=22: DX prob.}
\end{minipage} &
\begin{minipage}[c]{0.31\textwidth}
\centering
\includegraphics[width=\linewidth]{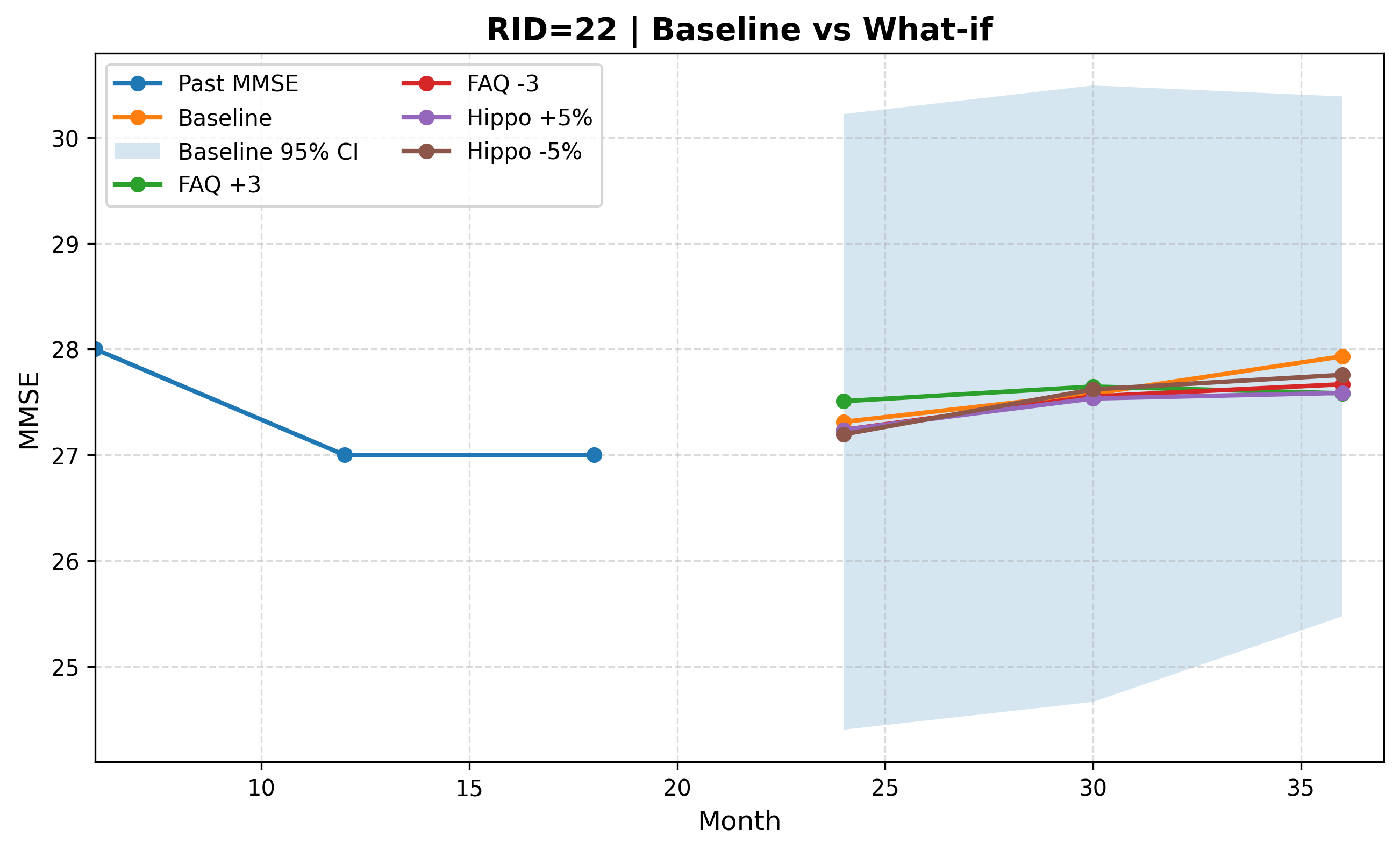}\\[-0.2em]
{\small (c) RID=22: what-if}
\end{minipage}\\[0.5em]

\begin{minipage}[c]{0.31\textwidth}
\centering
\includegraphics[width=\linewidth]{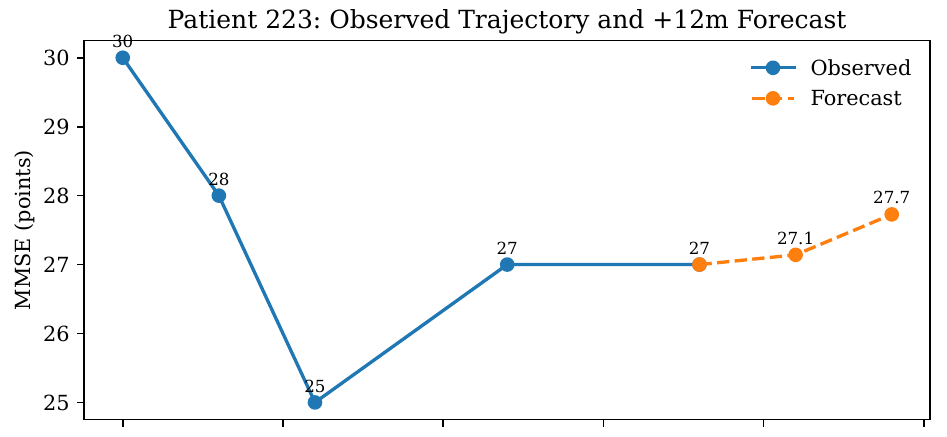}\\[-0.2em]
{\small (d) RID=223: MMSE}
\end{minipage} &
\begin{minipage}[c]{0.31\textwidth}
\centering
\includegraphics[width=\linewidth]{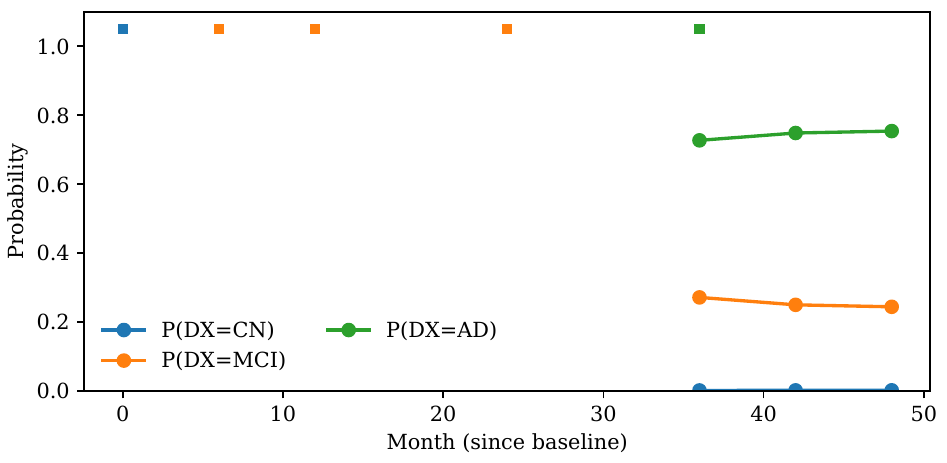}\\[-0.2em]
{\small (e) RID=223: DX prob.}
\end{minipage} &
\begin{minipage}[c]{0.31\textwidth}
\centering
\includegraphics[width=\linewidth]{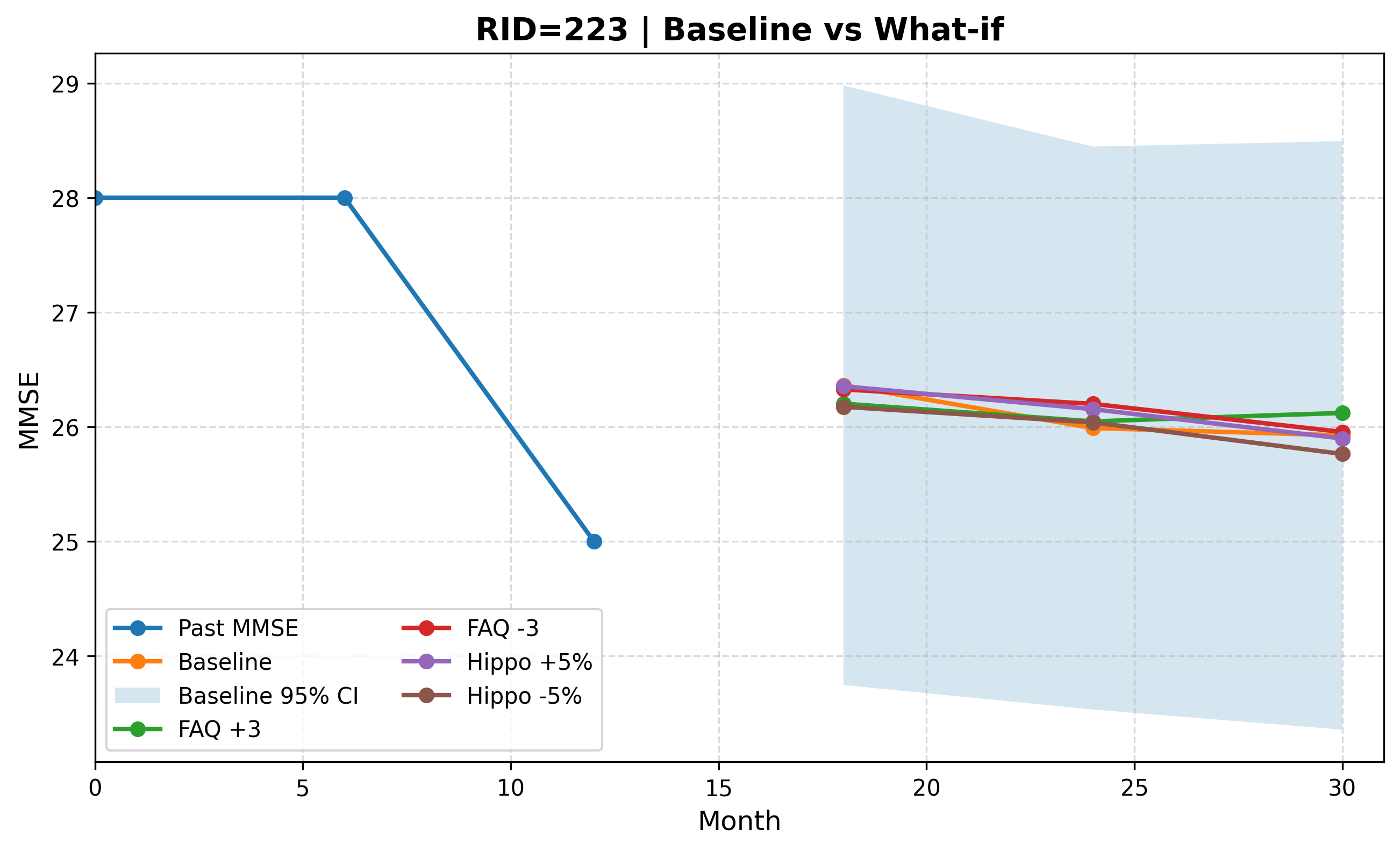}\\[-0.2em]
{\small (f) RID=223: what-if}
\end{minipage}\\[0.5em]

\begin{minipage}[c]{0.31\textwidth}
\centering
\includegraphics[width=\linewidth]{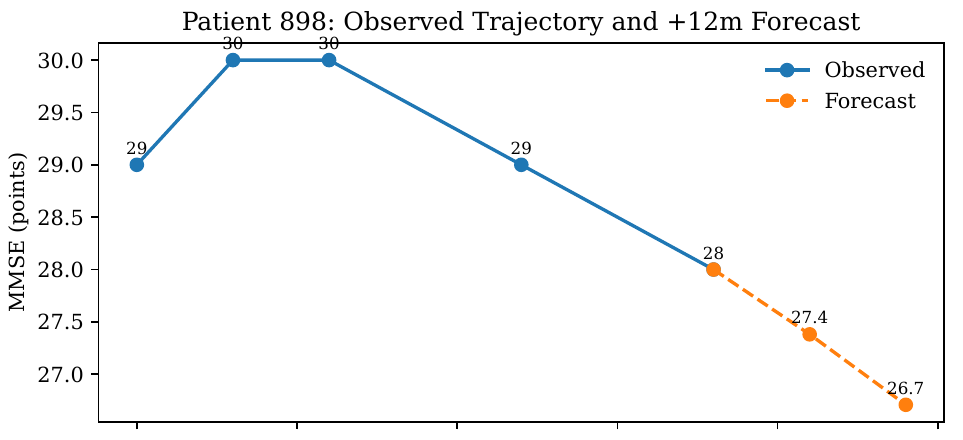}\\[-0.2em]
{\small (g) RID=898: MMSE}
\end{minipage} &
\begin{minipage}[c]{0.31\textwidth}
\centering
\includegraphics[width=\linewidth]{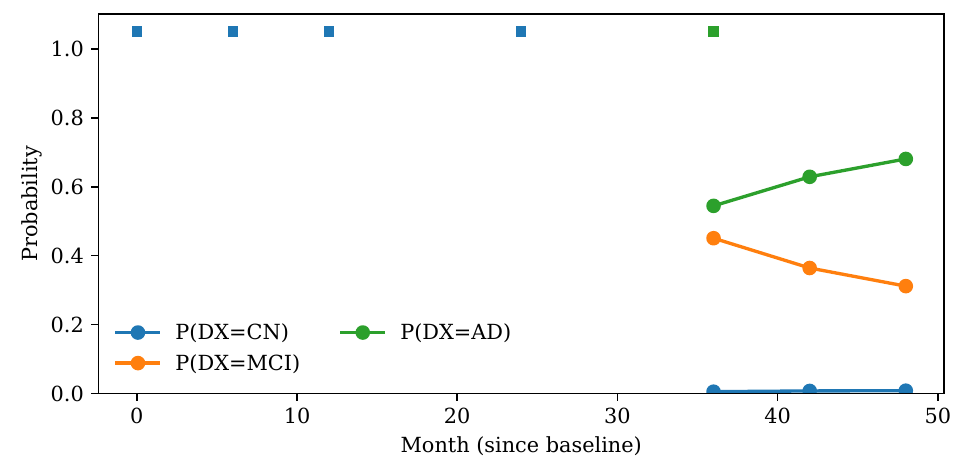}\\[-0.2em]
{\small (h) RID=898: DX prob.}
\end{minipage} &
\begin{minipage}[c]{0.31\textwidth}
\centering
\includegraphics[width=\linewidth]{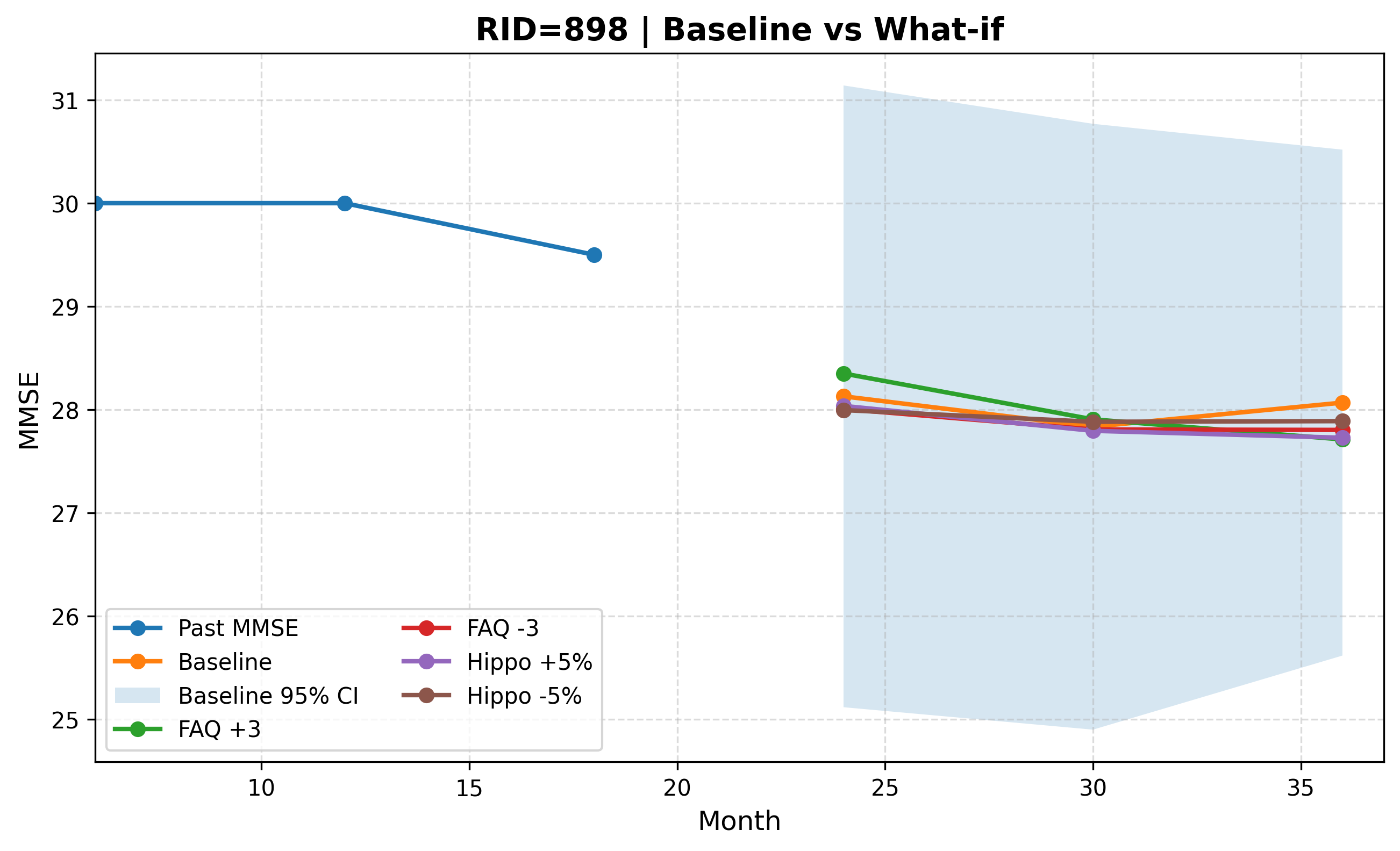}\\[-0.2em]
{\small (i) RID=898: what-if}
\end{minipage}

\end{tabular}
\caption{Representative patient-level digital-twin personas. Column 1: MMSE trajectory. Column 2: DX probability trajectory. Column 3: BiLSTM--Attention baseline vs what-if trajectories with 95\% predictive intervals.}
\label{fig:persona_grid}
\end{figure}

Among these, RID=223 serves as the illustrative example. As shown in the second-row panels of Fig. ~\ref{fig:persona_grid}, the digital twin captures a clinically relevant pattern in which MMSE remains close to 27 over the forecast horizon, while diagnosis probabilities continue to shift, with rising model-estimated AD risk and falling CN probability. This illustrates the practical value of the digital twin perspective: apparent cognitive stability on a single score coexists with model-estimated changes in diagnostic probabilities based on the broader multimodal profile. The MLP-based twin provides a subject-specific view anchored on the most recent observation, while the BiLSTM--Attention twin complements this by extrapolating the baseline-to-m18 history into a probabilistic m24--m36 trajectory with explicit uncertainty.

Representative \emph{what--if} simulations further show that FAQ perturbations of $\pm 3$ points and hippocampal perturbations of $\pm 5\%$ produce subtle but directionally consistent changes that remain close to the baseline forecast and mostly within the 95\% predictive interval. For this subject and forecast horizon, temporal uncertainty dominates over small covariate changes, but the framework still supports interpretable sensitivity analysis without retraining. The MLP- and BiLSTM-based twins therefore provide concordant subject-level interpretations, with differences in early forecast points remaining well explained by predictive uncertainty.

Fig.~\ref{fig:persona_grid} summarises the broader behaviour of the framework across all three personas. The first-row panels show RID=22 as a stable reference case, with narrow predictive intervals and minimal divergence under perturbation. The second-row panels show RID=223 as a more typical progressive monitoring scenario in which diagnosis risk evolves even while cognition appears relatively preserved. The third-row panels show RID=898 as a stable-appearing configuration in which apparently favourable cognition coexists with increasing model-estimated AD probability and wider uncertainty bands. Although these outputs are not prescriptive clinical recommendations, they demonstrate how the same digital twin framework can support individualised interpretation across stable, progressive, and latent-risk longitudinal contexts, while allowing newly observed visits to be assimilated for updated forecasts and uncertainty estimates over time.

\section{Discussion}
The strongest empirical result in this study is that the transition-based MLP outperformed the sequence-based BiLSTM-Attention branch for MMSE regression and diagnosis classification under the current data conditions. This finding is consistent with the different sample construction strategies used by the two branches. The MLP benefits from adjacent time-pair construction, which substantially enlarges the effective training set and aligns learning with local state transitions. By contrast, the temporal model must learn from fewer complete sequences and must extrapolate further into the future, making it more sensitive to irregular sampling, interpolation assumptions, and accumulated uncertainty.

The difference also reflects the nature of the signal being predicted. Structural MRI phenotypes evolve gradually and may provide a stable context, but MMSE changes between nearby visits are often small and noisy. In this setting, learning direct short-horizon transitions can be easier than learning higher-order longitudinal structure from sparse sequences. This helps explain why the simpler MLP is more effective as a predictive engine in the present dataset.

The main lesson is not simply that a smaller model performed better, but that the temporal formulation must match the structure and limitations of the clinical data. When longitudinal sampling is sparse and irregular, modelling adjacent state transitions can be more robust than attempting to learn full sequence dynamics. This has broader implications for healthcare AI, where dataset irregularity, limited follow-up, and missing observations are common. In such settings, practical robustness may be more clinically valuable than theoretical temporal expressiveness.

At the same time, the BiLSTM-Attention branch remains valuable because it extends the framework beyond point prediction to uncertainty-aware trajectory modelling. It produces individualised predictive intervals via Monte Carlo dropout and supports subject-level what-if simulation, which is central to the digital twin interpretation. In that sense, the two branches are not redundant: the MLP provides a strong short-term predictive baseline, while the temporal branch provides richer personalised trajectory analysis.

Our results suggest that model selection in longitudinal clinical prediction should be guided by data characteristics such as sparsity and irregular sampling, rather than increasing architectural complexity alone. In such settings, modelling local transition dynamics can be more robust than relying solely on sequence-based temporal representations, pointing toward a principled direction for future personalised modelling frameworks.

Several limitations should be noted. The study uses a single longitudinal cohort, sequence modelling depends on interpolation for irregular follow-up, and what-if analyses remain observational rather than causal. In addition, the MLP and BiLSTM branches differ in prediction horizon and effective sample construction, so their comparison should be interpreted as a practical comparison of temporal formulations rather than a strictly architecture-controlled benchmark. The current MLP and BiLSTM branches were intentionally kept relatively simple so that the comparison reflected differences in temporal formulation rather than heavy architecture engineering. 

A further limitation is that feature-level what-if perturbations provide model-based scenario exploration rather than causal intervention effects. The present framework is best interpreted as an uncertainty-aware forecasting and what-if tool rather than a causal simulator. Nevertheless, this type of analysis can be useful in longitudinal disease monitoring because it reveals how projected trajectories behave under small, clinically plausible perturbations. Such structured subject-level probing is difficult to obtain from standard endpoint classifiers and is one of the main practical strengths of the digital twin perspective. In addition, the evaluation is retrospective and limited to a single cohort, so external validation in independent datasets remains an important direction for future work.

\section{Conclusion}
This study presents a personalised digital twin framework for modelling Alzheimer’s disease progression from multimodal longitudinal data. The results show that model selection in clinical prediction tasks should be guided by data characteristics and task-specific constraints. While the framework achieved strong performance in multi-class diagnostic classification, transition-based MLP modelling was more effective for MMSE prediction under sparse and irregular follow-up conditions. More importantly, this work highlights a key insight for clinical machine learning: with real-world healthcare data characterised by limited samples, missing observations, and heterogeneous progression, greater model complexity does not necessarily lead to better performance. Instead, aligning the temporal modelling strategy with the structure of the data is critical. In such settings, modelling local transitions between adjacent visits can provide a more robust and data-efficient alternative to sequence-based approaches. Beyond predictive accuracy, the framework supports individualised trajectory estimation with uncertainty quantification and enables subject-specific what-if trajectory analysis. These capabilities offer a practical and interpretable approach to personalised monitoring and risk stratification in neurodegenerative disease contexts. Future work will focus on improving data-efficient modelling strategies, developing more principled methods for handling irregular longitudinal data, and extending the framework toward causal rather than observational scenario analysis. Validation across additional cohorts and incorporation of richer biomarker modalities will further strengthen clinical applicability. Ultimately, such advances may support more reliable and clinically individualised predictions for early detection and management of Alzheimer’s disease and related dementias.
\begin{credits}
\subsubsection{\ackname} This research was supported by the National Institute for Health and Care Research (NIHR). RA is supported by the NIHR Southampton Biomedical Research Centre. SM is supported by an NIHR Senior Clinical Practitioner Award.

\subsubsection{\discintname}
The authors have no competing interests to declare that are relevant to the content of this article.

\end{credits}
%
%
%
\bibliographystyle{splncs04}
\bibliography{references}

\begin{thebibliography}{10}
\providecommand{\url}[1]{\texttt{#1}}
\providecommand{\urlprefix}{URL }
\providecommand{\doi}[1]{https://doi.org/#1}

\bibitem{amato2025digital}
Amato, L.G., Lassi, M., Vergani, A.A., Carpaneto, J., Mazzeo, S., Moschini, V., Burali, R., Salvestrini, G., Fabbiani, C., Giacomucci, G., et~al.: Digital twins and non-invasive recordings enable early diagnosis of alzheimer's disease. Alzheimer's Research \& Therapy  \textbf{17}(1), ~125 (2025)

\bibitem{balasubramaniam2024machine}
Balasubramaniam, S., Sumina, S., Kumar, K.S., Prasanth, A.: Machine learning based models for implementing digital twins in healthcare industry. In: Metaverse Technologies in Healthcare, pp. 135--162. Elsevier (2024)

\bibitem{elgammal2025digital}
Elgammal, Z., Albrijawi, M.T., Alhajj, R.: Digital twins in healthcare: a review of {AI}-powered practical applications across health domains. Journal of Big Data  \textbf{12}(1),  1--28 (2025)

\bibitem{fekonja2024digital}
Fekonja, L.S., Schenk, R., Schr{"o}der, E., Tomasello, R., Tomsic, S., Picht, T.: The digital twin in neuroscience: from theory to tailored therapy. Frontiers in Neuroscience  \textbf{18},  1454856 (2024)

\bibitem{fisher2019machine}
Fisher, C.K., Smith, A.M., Walsh, J.R.: Machine learning for comprehensive forecasting of alzheimer's disease progression. Scientific Reports  \textbf{9}(1),  13622 (2019)

\bibitem{hasan2024explainable}
Hasan~Saif, F., Al-Andoli, M.N., Bejuri, W.M.Y.W.: Explainable {AI} for alzheimer detection: A review of current methods and applications. Applied Sciences  \textbf{14}(22),  10121 (2024)

\bibitem{karaman2024assessing}
Karaman, B.K., Sabuncu, M.R.: Assessing the significance of longitudinal data in alzheimer's disease forecasting. In: International Conference on AI in Healthcare. pp. 3--16. Springer (2024)

\bibitem{kumar2024himal}
Kumar, S., Yu, S.C., Michelson, A., Kannampallil, T., Payne, P.R.: {HiMAL}: Multimodal hierarchical multi-task auxiliary learning framework for predicting alzheimer's disease progression. JAMIA Open  \textbf{7}(3),  ooae087 (2024)

\bibitem{marinescu2020alzheimer}
Marinescu, R.V., Oxtoby, N.P., Young, A.L., Bron, E.E., Toga, A.W., Weiner, M.W., Barkhof, F., Fox, N.C., Eshaghi, A., Toni, T., et~al.: The alzheimer's disease prediction of longitudinal evolution ({TADPOLE}) challenge: results after 1 year follow-up. arXiv preprint arXiv:2002.03419  (2020)

\bibitem{norton2014potential}
Norton, S., Matthews, F.E., Barnes, D.E., Yaffe, K., Brayne, C.: Potential for primary prevention of alzheimer's disease: an analysis of population-based data. The Lancet Neurology  \textbf{13}(8),  788--794 (2014)

\bibitem{odusami2024machine}
Odusami, M., Maskeliunas, R., Damasevicius, R., Misra, S.: Machine learning with multimodal neuroimaging data to classify stages of alzheimer's disease: a systematic review and meta-analysis. Cognitive Neurodynamics  \textbf{18}(3),  775--794 (2024)

\bibitem{qiang2023diagnosis}
Qiang, Y.R., Zhang, S.W., Li, J.N., Li, Y., Zhou, Q.Y.: Diagnosis of alzheimer's disease by joining dual attention {CNN} and {MLP} based on structural {MRI}s, clinical and genetic data. Artificial Intelligence in Medicine  \textbf{145},  102678 (2023)

\bibitem{ren2025utilization}
Ren, Y., Pieper, A.A., Cheng, F.: Utilization of precision medicine digital twins for drug discovery in alzheimer's disease. Neurotherapeutics p. e00553 (2025)

\end{thebibliography}
\end{document}